\documentclass[sigconf]{acmart}

\pdfoutput=1
\AtBeginDocument{%
  \providecommand\BibTeX{{%
    \normalfont B\kern-0.5em{\scshape i\kern-0.25em b}\kern-0.8em\TeX}}}
    
\copyrightyear{2020} 
\acmYear{2020} 
\setcopyright{acmcopyright}\acmConference[CIKM '20]{Proceedings of the 29th ACM International Conference on Information and Knowledge Management}{October 19--23, 2020}{Virtual Event, Ireland}
\acmBooktitle{Proceedings of the 29th ACM International Conference on Information and Knowledge Management (CIKM '20), October 19--23, 2020, Virtual Event, Ireland}
\acmPrice{15.00}
\acmDOI{10.1145/3340531.3412740}
\acmISBN{978-1-4503-6859-9/20/10}
\newfont{\mycrnotice}{ptmr8t at 7pt}
\newfont{\myconfname}{ptmri8t at 7pt}

\usepackage{soul}
\usepackage{url}
\usepackage{fancyhdr}
\usepackage[utf8]{inputenc}
\usepackage{graphicx}
\usepackage{subcaption}
\usepackage{multirow}
\usepackage{amsmath}
\usepackage{amsthm}

\usepackage{amssymb}
\usepackage{booktabs}
\usepackage{algorithm}
\usepackage{algorithmic}
\usepackage{comment}
\usepackage{hyperref}

\begin{document}
\title{Query-aware Tip Generation for Vertical Search}

\author{Yang \ Yang$^1$ , Junmei \ Hao$^2$, Canjia \ Li$^3$, Zili \ Wang$^4$, Jingang Wang$^1$, Fuzheng Zhang$^1$,Rao Fu$^1$, Peixu Hou$^1$, Gong Zhang$^1$, and Zhongyuan Wang$^1$}

\authornote{Yang Yang, Junmei Hao, Canjia Li and Zili Wang are equal contributions, and Jingang Wang is the corresponding author.}

\affiliation{
\textsuperscript{1}Meituan\quad 
\textsuperscript{2}Xi'an Jiaotong University\quad
\textsuperscript{3}University of Chinese Academy of Sciences\quad
\textsuperscript{4}Xidian University\\
\{yangyang113, wangjingang02, zhangfuzheng, furao02,wangzhongyuan02\}@meituan.com, \\ \{peixu.hou,
gong.zhang\}@dianping.com, \\ haojunmei1996@stu.xjtu.edu.cn,
licanjia17@mails.ucas.ac.cn, ziliwang.do@gmail.com}

 \begin{abstract}
As a concise form of user reviews, tips have unique advantages to explain the search results, assist users' decision making, and further improve user experience in vertical search scenarios.
Existing work on tip generation does not take query into consideration, which limits the impact of tips in search scenarios.
To address this issue, this paper proposes a query-aware tip generation framework, integrating query information into encoding and subsequent decoding processes.
Two specific adaptations of Transformer and Recurrent Neural Network (RNN) are proposed.
For Transformer, the query impact is incorporated into the self-attention computation of both the encoder and the decoder.
As for RNN, the query-aware encoder adopts a selective network to distill query-relevant information from the review, while the query-aware decoder integrates the query information into the attention computation during decoding.
The framework consistently outperforms the competing methods on both public and real-world industrial datasets.
Last but not least, online deployment experiments on Dianping demonstrate the advantage of the proposed framework for tip generation as well as its online business values.
\end{abstract}


\maketitle

\section{Introduction}


In generic web search, given a user query, the search engines return a list of relevant documents and usually present title-snippet pairs to users. 
While in vertical searches, such as Yelp\footnote{\url{https://www.yelp.com/}} and Dianping\footnote{\url{https://www.dianping.com/}, a leading vertical service platform in China, whose content including food, entertainment, and travel services, etc.}, to better assist users' decision making, the vertical search engine usually presents some information in addition to search results. 
\begin{figure}[htb]
	\centering
	\begin{subfigure}[t]{0.45\columnwidth}
		\includegraphics[width=\textwidth]{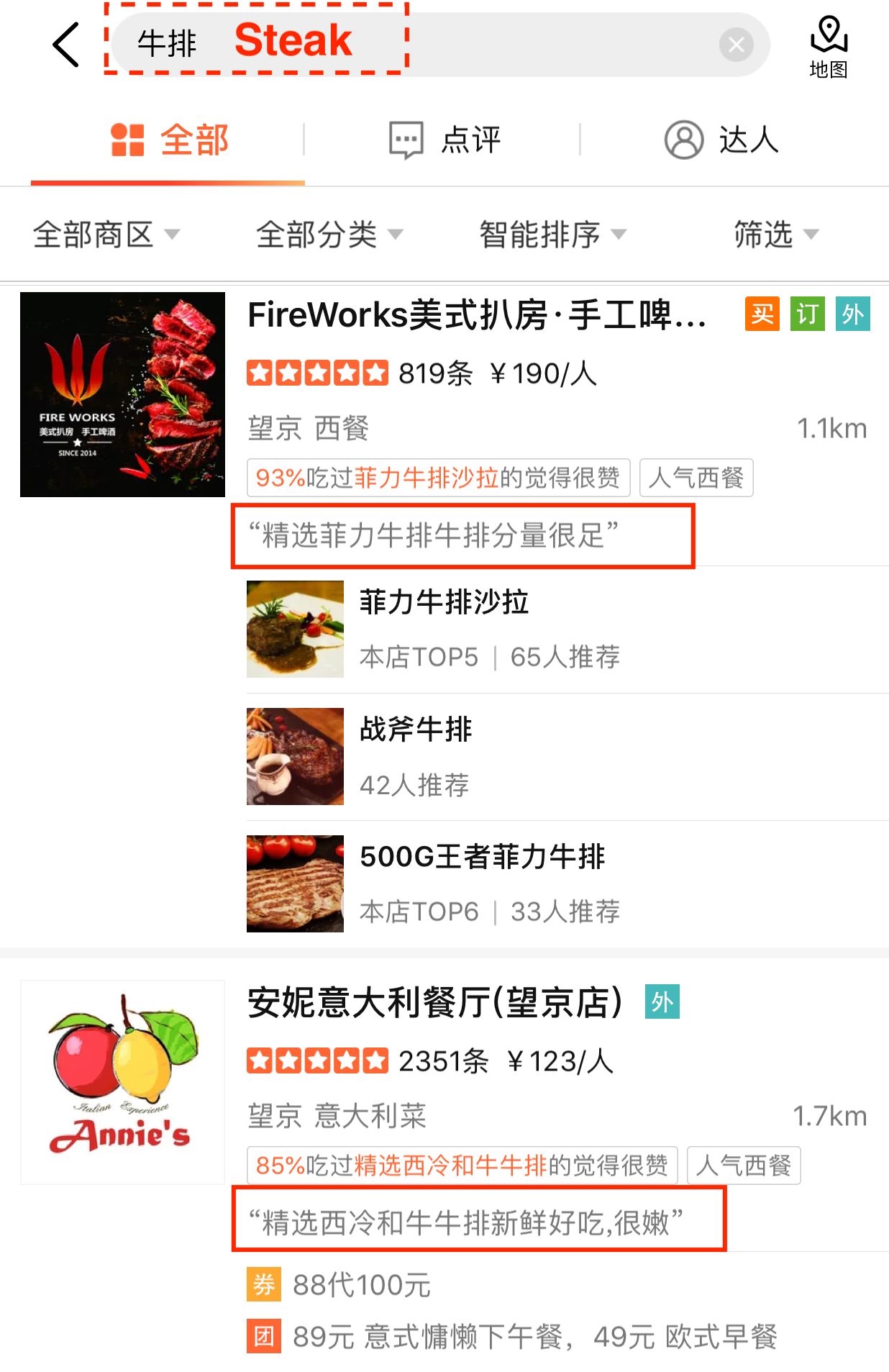}
		\caption{\small{Search result page of ``Steak''. The upper tip is ``\textit{The selected filet steak is enough served.}'', and the bottom one is ``\textit{The selected Japanese sirloin steak is fresh, good taste and tender.}''.} }
		\label{subfig:srp}
	\end{subfigure}
	\hspace{1em}
	\begin{subfigure}[t]{0.45\columnwidth}
		\includegraphics[width=\textwidth]{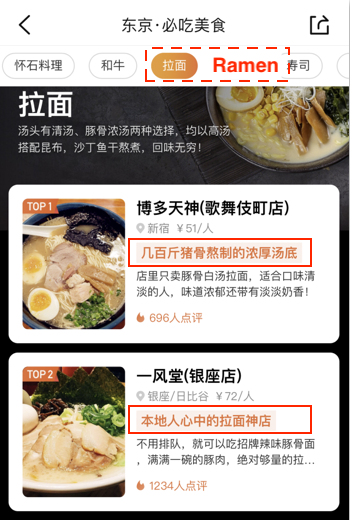}
		\caption{\small{An selected channel about ``Ramen''.  The upper tip is ``\textit{Pleasant smell and tasty soup made from pork bones}'', and the bottom one is ``\textit{The best Ramen rated by natives.}''.}}
		\label{subfig:ramen}
	\end{subfigure}
	\caption{Two vertical search scenarios on Dianping App. The queries are boxed with dash lines and the tips are boxed with solid lines.}
\vspace{-3pt}
	\label{fig:dianping}
\end{figure}

\autoref{fig:dianping} demonstrates two typical vertical search scenarios on Dianping, a popular E-commerce application in China.
\autoref{subfig:srp} presents the top $2$ restaurants (also known as place of interests (POI)) in the search result page (SRP) of the query ``Steak''.
The SRP not only presents essential information such as price and user ratings of returned restaurants but also provides some one-sentence tips (boxed with red solid lines). 
\autoref{subfig:ramen} shows a selected channel with tips after users clicking the topic ``Ramen''.
The displaying topics can be treated as potential user queries.
These tips, usually as compact and concise feature highlights of the listed POIs, are especially valuable for users to get a quick insight over the search results. 
Moreover, tips can be utilized to provide fine-grained and more reliable search explanations to help consumers making more informed decisions.

Obviously, it is impractical to manually write tips for millions of POIs indexed by vertical search platforms.
Fortunately, large amounts of user-generated reviews have been accumulated for these POIs. Hence it is intuitive to distill relevant information from reviews as tips.
Based on user reviews, earlier work generate tips for POIs with natural language generation techniques, such as unsupervised extractive~\cite{weber2012answers,zhu2018unsupervised} and abstractive methods~\cite{li2017neural,li2019neural}. 
As effective as they are, these tips are to some extent sub-optimal as they are generated without taking the user queries into consideration.

This motivates us to focus on producing tips by harnessing both the user query and the POI's reviews.
Such query-aware tips can potentially answer user's intent and attract users' attention better than a query-agnostic alternative.
For example, given a query ``Coffee Latte'',  a tip ``The vanilla latte tastes great!'' is more informative than a tip ``I love the bubble tea.'' from the view of user experience.

To this end, we propose query-aware tip generation for vertical search. 
There are two popular architectures for encoder-decoder framework, {\it i.e.}, Transformer \cite{attentionisalluneed} and RNN ~\cite{D14-1179,sutskever2014sequence}.
Accordingly, we develop query-aware tip generation encoders and decoders based on them, respectively.

{\bf Query-aware Encoder (\textsc{Qa\_Enc})}.
	For Transformer, we incorporate the query representation into the self-attention computation.
	For RNN, we introduce a selective gate network in the encoder to distill query-relevant information from the input sequence.
	
{\bf Query-aware Decoder (\textsc{Qa\_Dec})}.
	For Transformer, we similarly incorporate the query representation into the self-attention computation as the encoder.
	For RNN, we improve the attention mechanism by integrating query representation into the context vector to better direct the decoder.

To the best of our knowledge, this is the first work focusing on query-aware tip generation for vertical e-commerce search. 
The main contributions can be summarized as follows:
\begin{itemize}
	\item We propose a query-aware tip generation framework, which is intuitive but effective in vertical search scenarios.
	\item We introduce query-aware encoders and decoders to enhance the encoder-decoder framework to produce query-aware tips from user reviews, based on Transformer and RNN.
	\item We evaluate our framework on both public and real-world industrial datasets. 
	Extensive experimental results indicate the effectiveness of our framework. 
	We have also deployed our method in a real-world e-commerce platform and observed better performance than the competing baseline models.
\end{itemize}
\section{Query-aware Tip Generation Framework}
This section introduces the proposed query-aware encoder-decoder framework in detail. 
Similar to seq2seq text generation, the objective of query-aware tip generation is to generate a concise tip given a piece of review, except that there exists auxiliary information, i.e., a user query. 
There are two popular neural network architectures for encoders and decoders, i.e., the Transformer and the RNN. 
Both of the two architectures are adapted to involve query information.
Specifically, the query can be utilized in the encoder and the decoder separately or jointly.

\subsection{Problem Formulation}
Given a user review $\mathcal{R}=\left(r_1, r_2, \cdots, r_N\right)$ of $N$ words, a tip generation system aims to generate a compact tip of length $M$, namely $\mathcal{T}=\left(t_1, t_2, \cdots, t_M\right)$, which is also relevant to user query $\mathcal{Q}=\left(q_1, q_2, \cdots, q_K\right)$ of $K$ query words.

\subsection{Transformer-based Adaptation}
Here we give a brief description of the more recent and arguably more superior Transformer text generation framework.
Basically, the Transformer model first projects the tokens in a sequence of length $n$ into the $d$-dimension embedding space, where these token embeddings are again projected into three different spaces, namely ${Q}$, ${K}$, ${V} \in \mathbb{R}^{n \times d}$, via three different projection matrices. Afterwards, the contextualized representations of the entire sequence are computed by multi-head scaled dot-product attention layer:
\begin{equation}
\begin{split}
    \text{Attention}(Q,K,V) & =  \text{softmax}(\frac{QK^T}{\sqrt{d}})V \\
    \text{head}_i & = \text{Attention}(QW^Q_i,KW^K_i,VW^V_i) \\
    \text{MultiHead}(Q,K,V) &  =  
    \text{Concat}(\text{head}_1,\ldots,\text{head}_h)W^O
\end{split}
\label{eq.dotproduct} 
\end{equation}
Where the projections are parameter matrices $W^Q_i \in \mathbb{R}^{d \times n}$, $W^K_i \in \mathbb{R}^{d \times n}$, $W^V_i \in \mathbb{R}^{d \times n}$ and $W^O \in \mathbb{R}^{hd \times d}$, and $h$ denotes the number of the attention heads.
In the Transformer, the ${Q}, {K}, {V}$ are from the embeddings of the same input sequence, hence the attention layer is revised to self-attention layer, which helps the Transformer to summarize other words of the input sentence at the current position.
The outputs of the self-attention layer are fed to the layer-normalization and position-wise feed-forward neural network. 
The Transformer encoder block can be stacked one by one to obtain the abstract representation of each token. 
When the Transformer is applied to the text generation task, the outputs of the last encoder block are taken as key and value weights for decoding.
Besides the self-attention and feed-forward layers, the decoder block is also equipped with an encoder-decoder attention layer in between to focus on the relevant parts of the input sequence.

The above mechanism is modified in our proposed Transformer-adapted framework, which consists of three components: (1) a review-aware query encoder, (2) a query-aware review encoder, and (3) a query-aware tip decoder.

\begin{figure}[!tb]
 	\centering
	\includegraphics[width=0.32\textwidth, height=0.40\textwidth]{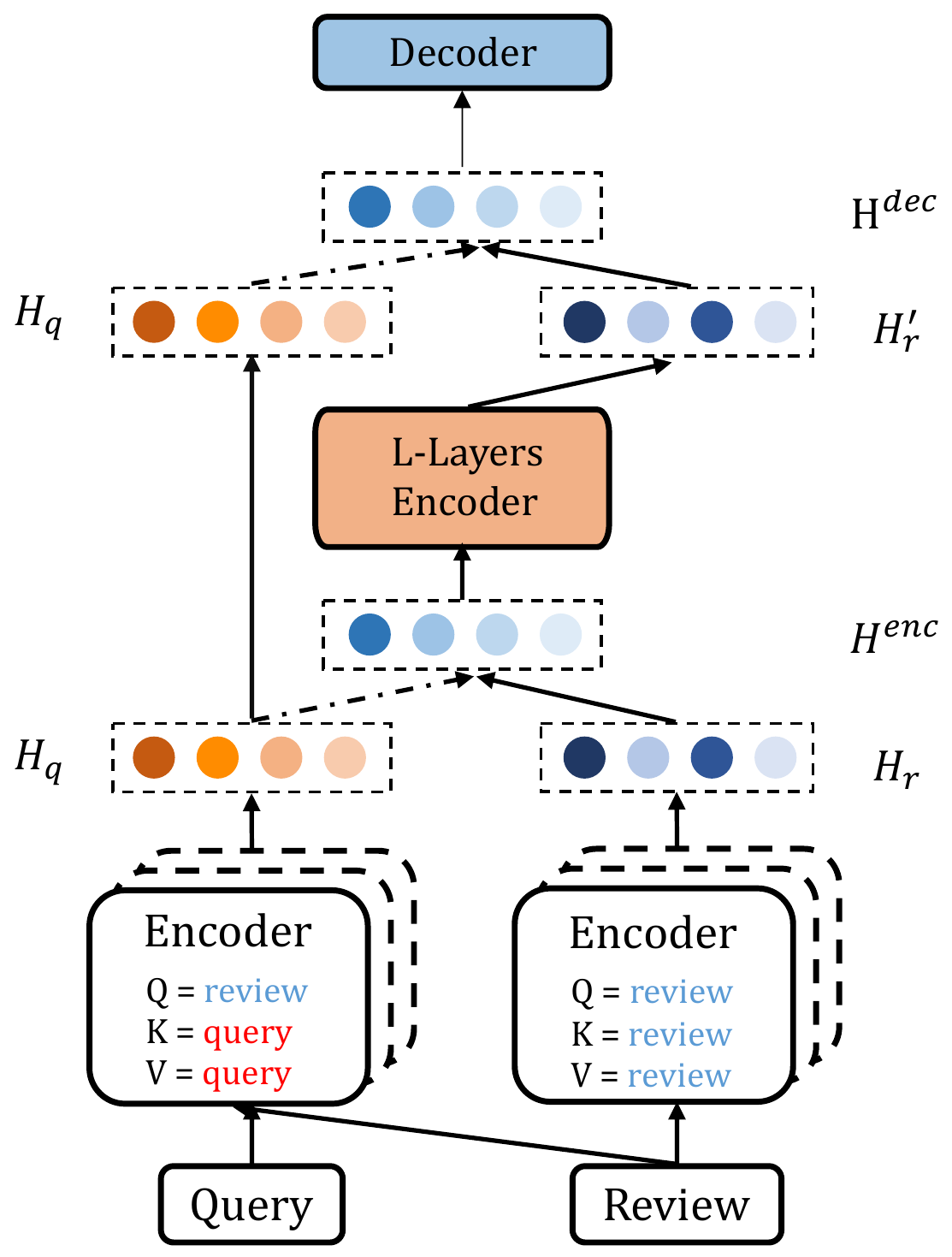}
	\caption{The Transformer-based query-aware tip generation framework. The left encoder encodes review-aware query while the right encoder maintains a self-attention manner to encode review. The two dotted arrows indicate the query information flows to the encoder and the decoder respectively.}
\vspace{-10pt}
 	\label{fig:Transformer}
\end{figure}

\vspace{-3pt}
\paragraph{Review-aware Query Encoder.}
The user query represents a condensed information need, while the review contains more detailed information for the POI.
The heterogeneity makes information sharing and matching difficult.
To bridge this semantic gap, we first introduce a review-aware query encoder to represent the query.
More concretely, the dot product attention layer (i.e., self-attention layer in Transformer) is adjusted to allow the interaction between the user query and the review.

Given a query $\mathcal{Q}$ and a review $\mathcal{R}$, we first use $E_q = \text{Emb}({\mathcal{Q}}) $ and $E_r = \text{Emb}({\mathcal{R}})$ to represent their word embeddings. 
Then the review contextualized information was incorporated into query representation. 
Formally, the dot product attention layer in \autoref{eq.dotproduct} can be modified as:
   $ H = \text{MultiHead}(E_r,{E_q},{E_q})$
$H \in \mathbb{R}^{N \times d}$ is then fed into the feed-forward and layer normalization layers as in the Transformer. The output of the query encoder is represented as $H_q \in \mathbb{R}^{N \times d}$. The output of the query encoder can be adapted into either review encoding or tip decoding in the tailored Transformer-based framework.

\vspace{-3pt}
\paragraph{Query-aware Review Encoder.} Similarly, the contextualized representations of the review $\mathcal{R}$ is obtained by setting $Q, K, V = E_r$. The output is denoted as $H_r \in \mathbb{R}^{N \times d}$.

Intuitively, $H_r$ encodes the user review's sequential representations. $H_q$ digests the query information under the context of the review. 
To combine these two complementary representations, a feed-forward network layer is adopted, namely, ${H^{\text{enc}}} = [{H_q}; {H_r}]W$, where $W \in \mathbb{R}^{2d \times d}, {H^{\text{enc}}}\in \mathbb{R}^{N \times d} $. Subsequently, ${H^{\text{enc}}}$ is fed into several Transformer encoder layers to further extract the query-aware representation for each token.

\vspace{-3pt}
\paragraph{Query-aware Tip Decoder.} 
When adapting query into tip decoding, the review and query are met at the early stage and composed by stacked Transformer encoder layers. 
To distinguish these two information flows and encourage the decoder to obtain the query message directly during decoding, stacked Transformer encoder layers are only applied to review hidden representation ${H_r}$ to obtain a deeper contextualized representation ${H_r'}$.
During decoding, ${H_r'}$ and ${H_q}$ are combined to generate the key and value matrices, namely, ${H^{dec}} = [H_q; {H_r'}]W$, where ${H^{dec}}\in \mathbb{R}^{N \times d}$.

The above encoding and decoding mechanisms are illustrated in \autoref{fig:Transformer}.
Please note that ${H^{dec}}$ is used as the key and value matrix for decoder layers during decoding.


\subsection{RNN-based Adaptation}
Besides the Transformer, another family of text generation model is the RNN-based encoder-decoder network.
Such models can be integrated into our tip generation framework conveniently.
The RNN-adapted framework also consists of three components, i.e., (1)a query encoder, (2)a query-aware review encoder, and (3)a query-aware tip decoder.

\begin{figure}[!t]
 	\centering
	\includegraphics[width=0.4\textwidth]{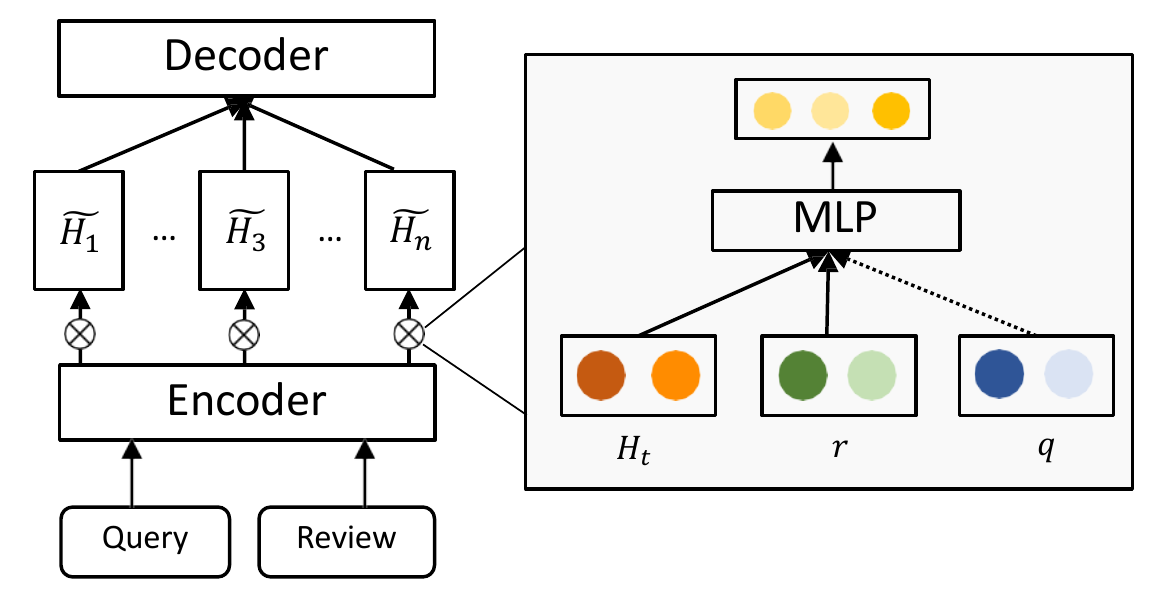}
	\caption{The RNN-based query-aware tip generation framework.}
 	\label{fig:rnn}
\end{figure}

\paragraph{Query Encoder.}
A Bidirectional Long Short-Term Memory (Bi-LSTM)~\cite{zhang-etal-2015-bidirectional} is adopted to generate a hidden representation $h_q \in \mathbb{R}^{2d}$ of the user query, where $h_q$ is the concatenation of the last hidden state of the forward pass and the first hidden state of the backward pass.

\paragraph{Query-aware Review Encoder.}
Similarly, another Bi-LSTM generates the hidden states ${H} \in \mathbb{R}^{N\times 2d}$ of all input review tokens.
Afterwards, the review hidden representation $h_{r} \in \mathbb{R}^{2d}$ is obtained in the same way as $h_q$.

At each time step $t$, to distill the query-relevant information from the review sequence, a selective gate~\cite{P17-1101}, denoted as $g_{t}$, is calculated as follows.
\begin{equation}
{g}_t=\sigma\left({W}_{r}\left[{H}_t;h_{r}\right]+{ W}_q{h_q} + {b}_g\right)
\end{equation}
where ${W}_{r} \in \mathbb{R}^{2d \times 4d}, {W}_q \in \mathbb{R}^{2d \times 2d}$ and ${b}_g\in \mathbb{R}^{2d}$ are learnable parameters, and $\sigma$ is sigmoid activation function. 
Thereby, the hidden representation of each time step $t$ is updated as 
\begin{equation}
\tilde{H}_{t} = {g}_t \circ {H}_{t} \label{eq:new_encode_output}
\end{equation}
where $\circ$ is an element-wise multiplication.
Such a gate acts like a ``soft'' selector selecting those tokens in the review that are relevant to the query.

\paragraph{Query-aware Tip Decoder.}
During decoding, the unidirectional LSTM decoder receives the output token from last step and has a hidden state ${s}_t$.
To encourage the decoder to generate a word that is relevant to the query, an intuitive but effective approach is to introduce the query representation to guide the decoding process, i.e., the attention ${a}_t$ over different input tokens. 
To implement it, at each decoding step $t$, we have:
\begin{eqnarray}
c_i & = & W_c[\tilde{H}_i;s_t] + W_qh_q + b_c \\
a_i^t & = & \text{softmax}\left( v^T \tanh(c_i)\right)  \label{eq:attention_renew}
\end{eqnarray}
where ${W}_q \in \mathbb{R}^{d \times 2d}, W_c \in \mathbb{R}^{d \times 3d}, \mathbf{b}_c \in \mathbb{R}^d, \mathbf{v} \in \mathbb{R}^d$.
The attention distribution ${a}^t$ varies at each time step, assuring the model to leverage the query and decoder hidden state into the contextual vector computation.
The resulting time-dependent encoder hidden state is calculated as the following:
\begin{equation}
    {h}_t^\text{dec} =  \sum_i a_i^t \tilde{H}_{i} 
\end{equation}
which is used for subsequent decoding.
The above encoding and decoding mechanisms are illustrated in \autoref{fig:rnn}.

\subsection{Model Training}
At each time step $t$, the output of the decoder is denoted as $h_t^d \in \mathbb{R}^{d}$.
The soft-max function is adopted to normalize the distribution.
\begin{equation}
P_t= \text{softmax}({W}_v{h}_t^d)
\end{equation}
where ${W}_v \in \mathbb{R}^{V \times d}$ maps the hidden state into a $V$-dimension vocabulary space.
${P_t}$  provides a final normalized distribution for token prediction.
During training, the negative log likelihood loss for each training sample is defined as follows:
\begin{equation} \label{eq.loss}
    \mathcal{L} = - \frac{1}{M} \sum_{t=1}^{M} {P_t}(y_t)
\end{equation}
where $y_t$ is the index of ground-truth. 
\section{Datasets}
To evaluate the effectiveness of the proposed query-aware encoders and decoders thoroughly, we conduct an extensive set of experiments on two datasets, \textsc{Debate} and \textsc{Dianping}.
\subsection{Debate}
The first dataset used is an open-source query-based English summarization dataset~\cite{nema2017diversity}, denoted as \textbf{\textsc{Debate}} for simplicity.
The dataset is created from Debatepedia\footnote{\url{http://www.debatepedia.org/en/index.php/Welcome_to_Debatepedia\%21}}, an encyclopedia of pro and con arguments and quotes on critical debate topics. 
The queries associated with the topic, the set of documents and an abstractive summary associated with each query which is not extracted directly from the document are crawled from Debatepedia. 
The dataset\footnote{\url{https://github.com/PrekshaNema25/DiverstiyBasedAttentionMechanism}} includes $13,719$ ({\it query, document, summary}) triplets in total\footnote{Note that the quantities of triplets released on Github is not consistent as depicted in ~\cite{nema2017diversity}}.
\autoref{tab:debate_example} presents an example.

\begin{table}[htbp]
\small
  \centering
  \caption{A (document, query, summary) example in \textsc{debate}.}
    \begin{tabular}{lp{0.35\textwidth}}
    \hline
    Document & The “natural death” alternative to euthanasia is not keeping someone alive via life support until they die on life support. That would, indeed, be unnatural. The natural alternative is, instead, to allow them to die off of life support. \\
    \hline
    Query & "non-treatment" : Is euthanasia better than withdrawing life support? \\
    \hline 
    Summary & The alternative to euthanasia is a natural death without life support. \\
    \hline
    \end{tabular}%
  \label{tab:debate_example}%
\end{table}%

\subsection{Dianping}
Due to the lack of public tip generation datasets for e-commerce search, we create an in-house dataset by crawling the search log of the Dianping App. 
Dianping is a leading Chinese vertical E-commerce platform where customers can write reviews for POIs such as restaurants, hotels, etc.
The dataset is denoted as \textbf{\textsc{Dianping}} for simplicity.
Currently, existing tips can be categorized as manually-extractive, manually-abstractive, and template-based (e.g., \textit{The best [placeholder] in town.}). 
As their names suggest, these tips are created by human experts or manual rules, normally based on concrete reviews, and organized in a one (POI) to many (tips) manner, respectively. 
These tips can be presented in various scenarios.
For instance, for query-agnostic recommendation, when a POI is recommended to the user, one of its tips is randomly selected.
In scenarios involving queries, to enhance the user experience, Dianping selects a tip that is the most similar to the query, as shown in~\autoref{fig:dianping}. We collected query logs with such query-aware tips for our experiments. 

For a query-aware tip, we associate it with its review as follows.
If a tip is created from a specific review, the original review is simply retrieved.
For tips such as template-based ones, or those whose original review is unidentifiable, we would use the tip itself to retrieve the most relevant review from the reviews related to the POI. 
The relevance score is calculated with the BM25 weighting method~\cite{robertson2009probabilistic}.
If the relevance score is lower than a pre-defined threshold, which means there does not exist a potential informative review to produce the tip, the tip is removed from our dataset.
To facilitate the training process, some additional pre-processing is performed. 
Only reviews containing $100 \sim 150$ Chinese characters are kept. 
The tips containing more than $15$ characters are also filtered out,  as they cannot be displayed properly on a mobile phone screen. 

\begin{figure*}[htb]
	\centering
	\includegraphics[width=0.8\textwidth]{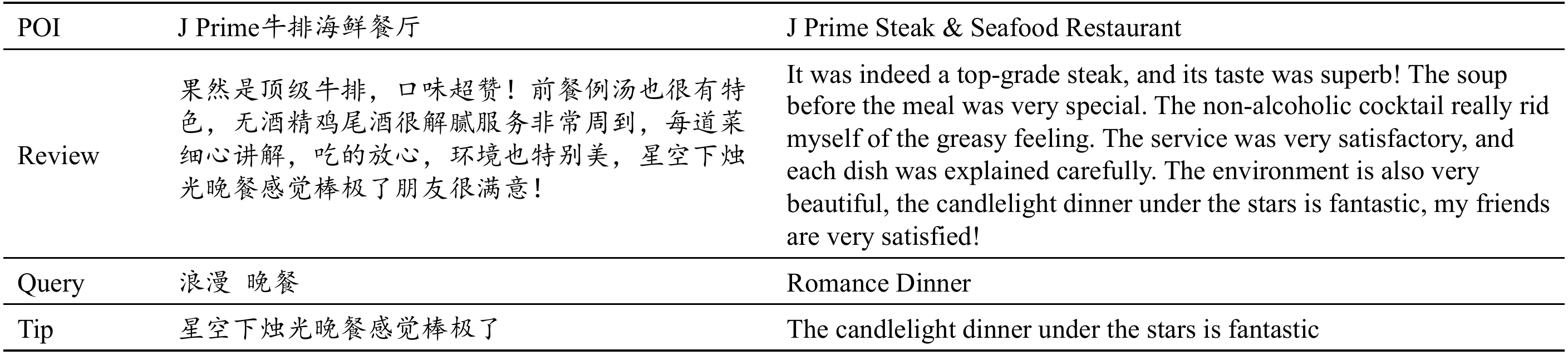}
	\caption{A (POI, review, query, tip) tuple example in our corpus. The column $3$ is the translation of the column $2$.}\label{fig:tuple}
\end{figure*}

We processed the search log spanning from July 1st to August 1st, $2019$. 
Finally, the corpus is composed of $224, 730$ ({\it POI, Review, Query, Tip}) tuples and the last three fields are used in our experiments. 
\autoref{fig:tuple} presents an example. Note that the words ``\textit{candlelight dinner}'' and ``\textit{stars}'' in the tip are all relevant to the query word ``\textit{romance}'' in Chinese culture.

The statistics of datasets are listed in table~\ref{tab:data_set}.
These tuples of both datasets are further randomly split into approximately $80\%$ for training, $10\%$ for validation and $10\%$ for testing. 
Note that the queries (questions) in \textsc{Debate} and \textsc{Dianping} vary   a lot. 
Moreover, to demonstrate the necessity of queries, two baselines without regard to queries are established in experiments. 
Hence, for these baselines, we remove queries from both datasets and then remove duplicates, resulting in their query-agnostic counterparts. 


\begin{table*}[htbp]
\small
  \centering
  \caption{Statistics of both datasets. The average length are calculated by words in English and characters in Chinese respectively. `w/o' represents without query while `w' means with query.}
    \begin{tabular}{lccccccc}
    \hline
    \multirow{2}{*}{Dataset} & \multicolumn{3}{c}{Avg\_Len} & \multirow{2}{*}{Query} & \multirow{2}{*}{Train} & \multirow{2}{*}{Valid} & \multirow{2}{*}{Test} \\ \cline{2-4}
    & Review (Doc) & Query & Tip (Summary) & & & & \\
    \hline
    \multirow{2}{*}{\textsc{Debate}}& \multirow{2}{*}{72.61}& \multirow{2}{*}{11.54} & \multirow{2}{*}{9.93}& w/o & $10,846$ &$1,356$ &$1,356$  \\
    \cline{5-8}
    & & && w/ &$10,975$ &$1,372$ & $1,372$ \\
    \hline
    \multirow{2}{*}{\textsc{Dianping}} & \multirow{2}{*}{101.50} & \multirow{2}{*}{3.39} & \multirow{2}{*}{12.20}& w/o & $137, 208$ & $17, 151$ & $17, 151$ \\
    \cline{5-8}
    & & && w/ & $179, 784 $ & $22, 473$ & $22, 473$ \\
    \hline
    \end{tabular}%
  \label{tab:data_set}%
\end{table*}%

\section{Experiments}
\subsection{Implementation Details}
In Transformer-based framework, all the Transformer variants are implemented as $6$ identical layers with a hidden size of $512$.
In RNN-based framework, both the encoder and decoder are implemented as $1$ layer of (Bi-)LSTM with a hidden size of $256$, and the word embedding size is set as $128$.
The word embedding is randomly initialized and learned during training. 
For optimization, we use Adam~\cite{adam} with initial learning rate $0.001$ and the batch size is empirically set as $128$. 
For \textsc{Dianping} dataset, the maximum encoding lengths for a review and a query are $150$ and $5$, respectively. Due to the limited screen size of mobile-devices, the maximum length of decoded tip is set as $15$. For the \textsc{Debate} dataset, the maximum encoding lengths for a review and a query are $160$ and $30$, and the maximum length of decoded tip is set as $30$. 

\begin{table*}[htbp]
\small
  \centering
  \setlength{\tabcolsep}{3pt}
  \caption{Automatic Evaluation Results on \textsc{Dianping} and \textbf{\textsc{Debate}} datasets.}
  \label{tb:exp-result}
    \begin{tabular}{llcccccc}
    \hline
    \multirow{2}{*}{Group} &
    \multirow{2}{*}{Methods} & \multicolumn{3}{c}{\textsc{Debate}} & \multicolumn{3}{c}{\textsc{Dianping}}  \\
    \cline{3-8} 
     &  &  Semantic & Lexicon & BLEU  &  Semantic & Lexicon & BLEU \\
    \hline
    \multirow{3}{*}{\textsc{Retrieval}} & \textsc{Query\_LEAD} & - & 10.23 & 2.23 & - & 40.70 & 23.20\\ 
    & \textsc{Extract\_{BM25}} & - & 14.39& 1.12 & - & 47.18 & 27.59 \\ 
    & \textsc{Extract\_{Embed}} & - & \bf{14.43}& 1.13 & - & 37.04 & 28.29 \\ 
    \hline
    \multirow{4}{*}{\textsc{RNN}} & \textsc{RNN}  & 83.87 & 8.91 & 11.02 & 60.08 & 40.94 & 40.74 \\
    & \textsc{RNN + Qa\_Enc} & 84.37& 9.23&  15.72& 62.65 & 41.37 & 48.29 \\
    & \textsc{RNN + Qa\_Dec} & 84.17& 9.07& 15.37& 62.77 & 43.92 & 46.88 \\
    & \textsc{RNN + Both} & 84.43& 9.34& 16.58& 64.86 & 44.11 & 48.38 \\
    \hline
    \multirow{4}{*}{\textsc{Transformer}} & \textsc{Trans} & 87.17& 10.52 & 30.41 & 65.64 & 47.49 & 48.71 \\
    & \textsc{Trans + Qa\_Enc} & 86.07& 13.17& 32.03& 67.00 & 49.79 & 50.39 \\
    & \textsc{Trans + Qa\_Dec}  & 84.70& 13.46& 32.52& 62.70  & 42.61 & 52.66 \\
    & \textsc{Trans + Both} & \bf{88.06}& 13.43& \bf{32.93}& \bf{69.79} & \bf{53.75} & \bf{54.20} \\
    \hline
    \end{tabular}%
  \label{tab:auto_evaluation}%
\end{table*}%

\subsection{Comparison Models}
Several competitive models are implemented to evaluate the performance of our query-aware tip generation framework.
Please note that both query and review are taken as model inputs.

{\bf\textsc{Query\_LEAD.}} Taking the leading sentence(s) of a document is reported to be a strong baseline in summarization~\cite{pointer-generator}. 
Here, 
the first sentence that contains the query in a review is extracted as the tip. 
If such a sentence can not be found, the leading sentence of the review is selected instead.

{\bf\textsc{Extract\_BM25.}} This is an unsupervised extractive baseline.
Given the query, the sentences in the review are ranked by their BM25 scores and the top one is favored.

{\bf\textsc{Extract\_Embed.}} This is another unsupervised extractive baseline. 
The sentences in the review are ranked by their embedding-based cosine similarities with the query. 
We use the publicly largest pre-trained Chinese word embedddings~\cite{song2018directional} for \textbf{\textsc{Ours}} and Glove\footnote{\url{http://nlp.stanford.edu/projects/glove/}} for \textbf{\textsc{Debate}}. 

{\bf\textsc{RNN.}} An abstractive baseline utilizing the pointer generator implementation, regardless of the query. 

{\bf\textsc{Trans.}} An abstractive baseline utilizing the Transformer-based encoder-decoder implementation, regardless of the query.


{\bf\textsc{RNN(Trans) + Qa\_Enc/Qa\_Dec/Both.}} The \textsc{RNN(Trans)} with the proposed \textsc{Qa\_Enc} and \textsc{Qa\_Dec} separately or jointly.

\subsection{Automatic Evaluation}
In automatic evaluation, the generated tips are assessed in terms of query-relevance and coherency. 

\underline{Metrics.} For query-relevance, two metrics are used. First, the cosine similarities between the embeddings of the generated tips and the corresponding queries are calculated, denoted as \textbf{Semantic}~\cite{average_and_extrema}.
The embedding of tip or query is obtained by maximizing over the embeddings of the tokens in the sequence.
Second, the number of co-occurring tokens in the generated tip and query divided by the query length is used as a lexical proxy of the relevance. 
This metric is denoted as \textbf{Lexicon}. 
The coherency is measured by \textbf{BLEU}~\cite{BLEU}. 

\underline{Results.} The results are reported in \autoref{tb:exp-result}. 
For the ease of presentation, all the metrics are multiplied by $100$. In general, the proposed method \textsc{Trans + Both} outperforms all the comparing algorithms in terms of almost all the metrics across both datasets.
In comparison to the $3$ query-aware retrieval-based baselines, the query-aware abstractive models perform almost better on both query-relevance and coherency criteria due to the flexibility of abstractive models.
Even the query is not explicitly mentioned in the review, the query-aware abstractive models can generate fluent tip relevant to the query.
In terms of \textbf{Lexicon} metric, the semantic-based retrieval method outperforms the other abstractive methods on the \textsc{Debate}.
We speculate it is caused by the long-query characteristic of the \textsc{Debate} dataset.
Recall that the queries in \textsc{Debate} are long questions in essence and \textbf{Lexicon} measures lexical similarity between the query and the final tip.
Retrieval-based methods focus on literal matching between the query and the extractive tip, while abstractive methods dedicate to generating fluent tips by attending to the key information in the given query.
In addition, Transformer-based models outperform RNN-based models in terms of all the metrics across both datasets. 
This is reasonable considering Transformer is better at handling the long-range dependencies in user reviews. 
For RNN-based and Transformer-based models, incorporating query-aware information including \textsc{Qa\_Enc} and \textsc{Qa\_Dec} improve the overall performance compared with the vanilla RNN and Transformer, which further verifies the importance of query-aware information.

\subsection{Manual Evaluation}
Due to the high cost of manual assessments, we only conduct manual evaluation of the proposed framework on \textsc{Dianping} dataset.
In manual evaluation, the generated tips of different models are assigned to $5$ annotators with a related background. 
They are instructed to score each generated tips with respect to $3$ perspectives: Readability, Relevance and Usefulness.
For Usefulness and Relevance, the majority annotating result is adopted as the final assessment, while for Readability the average annotating result is adopted.

\underline{Metrics.} Among the $3$ metrics, {\bf Readability} measures whether a generated tip is fluent and grammatical, {\bf Relevance} indicates whether a generated tip is relevant to the query, and {\bf Usefulness} demonstrates whether the generate tip is helpful for the user to make a decision.
In particular, Relevance and Usefulness are assessed by a binary score (i.e, $1$ for true and $0$ for false), and Readability is assessed by a $3$-point scale score from $1$ (worst) to $3$ (best). 

\underline{Results.} The results are reported in \autoref{tb:human-result}. 
Overall, the tips generated by Transformer-based models achieve better readability and query-relevance than RNN-based models. 
The proposed method \textsc{Trans + Both} performs best on all the metrics.

The introduction of query information into RNN and Transformer improves the relevance performance in both cases.
In terms of Usefulness, all the query-aware variants generate tips that are more informative for the users.


\begin{table}[htbp]
	\small
		\centering
		\caption{Manual Evaluation Results on \textsc{Dianping} dataset.
		}\label{tb:human-result}
		\begin{tabular}{llccc} \hline 
			\multicolumn{2}{c}{Methods}  & Read. & Rel. & Useful. \\ \hline 
			\multirow{4}{*}{RNN} & \textsc{RNN}   & 2.12 & 32.00\% & 43.56\% \\ 
			& \textsc{RNN + Qa\_Enc}  & 2.67 &  40.33\% & 43.13\% \\
			& \textsc{RNN + Qa\_Dec}  & 2.71 &  52.31\% & 41.49\% \\
			& \textsc{RNN + Both}  & 2.63 &  53.50\% & 44.50\% \\ \hline
			\multirow{4}{*}{Transformer} & \textsc{Trans}   & 2.55 & 39.25\% & 44.38\% \\
			& \textsc{Trans + Qa\_Enc}  & {\bf 2.88} & 60.52\% &47.36\% \\
			& \textsc{Trans + Qa\_Dec} & 2.80 & 60.53\% & 47.37\% \\
			& \textsc{Trans + Both}  & {\bf 2.88} &  {\bf 63.72\%} & {\bf 54.35\%} \\
			\hline 
		\end{tabular}
\end{table}

\subsection{Case Studies}
\begin{figure*}[htb]
 	\centering
	\includegraphics[width=0.86\textwidth]{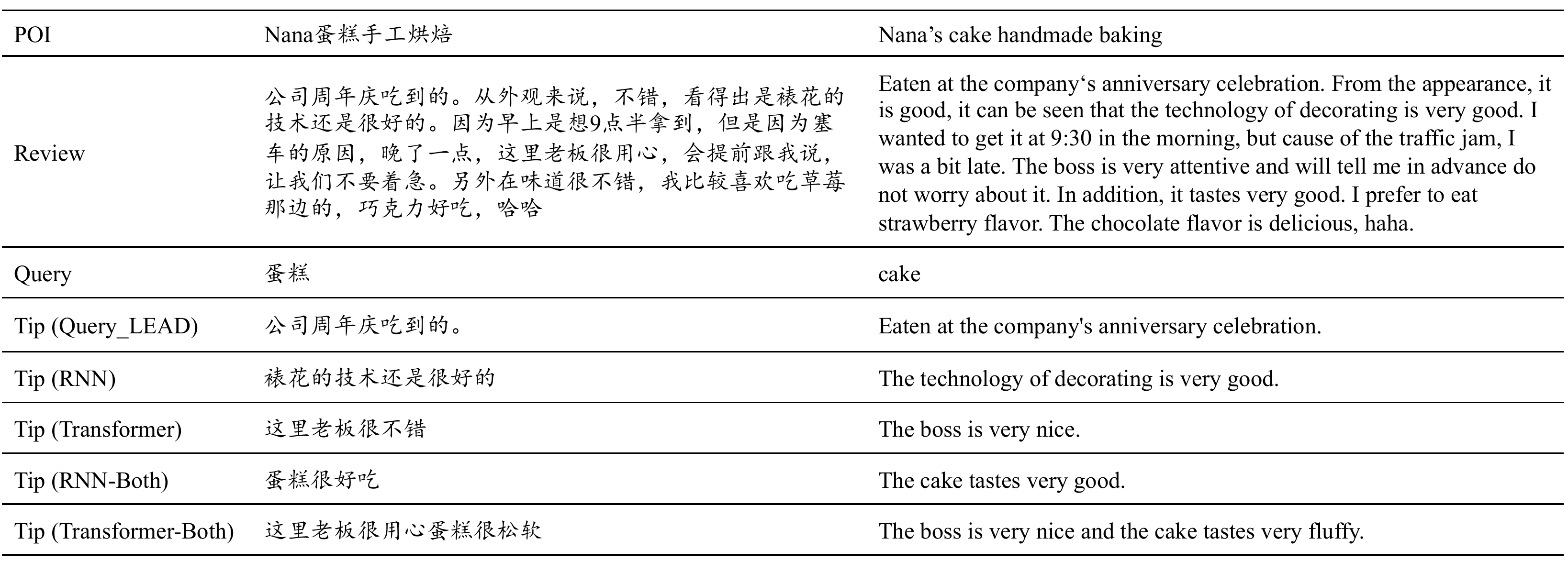}
	\caption{Examples of tip generation from Dianping dataset. The column $3$ is the translation of column $2$.}
 	\label{fig:case_study}
\end{figure*}

An illustrative case from the test set in \textsc{Dianping} dataset is presented in~\autoref{fig:case_study}. 
Due to the limited space, among query-aware models, only the results of \textsc{RNN + Both} and \textsc{Trans + Both} are presented.
It is obvious that only the two query-aware models generate tips related to the user query \textbf{cake}. 
What's more, the tip of \textsc{Trans + Both} mentions the shop owner's service attitude, which may be more informative to users.

\section{Online Deployment}
We deploy the query-aware tips in a production environment to test its online performance.
It is an A/B test in the SRP scenario (which is initiated by a user query) of the aforementioned App (with $\sim$10 million daily queries). 
It is noteworthy that the reviews belong to a POI are ranked by their number of ``likes'' by users.
Given a query, we take the top-ranked review of each returned POI to generate tips.
The A/B testing system diverts $10\%$ total query traffic and splits it equally into $4$ separate buckets. 
All the other settings of these buckets are identical.
The tips displayed in the $4$ buckets are generated with the following strategies: 
(1) No tip is displayed with POIs, (2) The tips are generated by \textsc{Trans}, (3) The tips are generated by \textsc{Extract\_BM25}, and (4) The tips are generated by \textsc{Trans + Both}.

The online test lasted for one week. 
CTR is adopted to test the performance, which is calculated as $ CTR=\frac{\#Clicks\_in\_SRP}{\#Query}$, where $\#Query$ is the count of the user queries, and $\#Clicks\_in\_SRP$ is the total clicks in the SRP triggered by the queries.
Higher CTR therefore implies that users are more likely to browse and click the POI.
Given a query, several clicks may occur in the same triggered SRP, $\#Clicks\_in\_SRP$ is counted as one in this case.
The averaged CTR in the $4$ buckets are $65.72\%$, $65.74\%$, $65.77\%$ and $65.80\%$, respectively.
In comparison to the No-tip baseline, even the query-agnostic tips improve the CTR. 
Both extractive and abstractive query-aware models (i.e., \textsc{Extract\_BM25} and \textsc{Trans + Both}) achieve higher CTR than query-agnostic \textsc{Trans}. 
The result is quite impressive, if we consider the fact that tips on a search result page occupy a relatively small space and thus only partially affect the users’ decision.
\section{Related Work}
\label{sec:related}
Our work touches on two strands of research within Query-focused text
summarization (QFS) and constrained sentence generation.

\noindent \textbf{Query-focused Summarization.} 
QFS aims to summarize a document cluster in response to a specific user query or topic.
It was first introduced in the Document Understanding Conference (DUC) shared tasks~\cite{dang2005overview,Li2007}.
Successful performance on the task benefits from a combination of IR and NLP capabilities, including passage retrieval and ranking, sentence compression~\cite{chali2012effectiveness,wang2016sentence}, and generation of fluent text.
Existing QFS work can be categorized into extractive and abstractive methods.
Extractive methods, where systems usually take as input a set of documents and select the top relevant sentences as the final summary. 
Cao et al.~\cite{C16-1053} propose AttSum to tackle extractive QFS, which learns query relevance and sentence saliency ranking jointly.
Abstractive methods attract more attention due to their flexibility in text summarization.
Rush et al.~\cite{D15-1044} first employ sequence-to-sequence (seq2seq) model~\cite{DBLP:conf/nips/SutskeverVL14} with attention mechanism~\cite{bahdanau+al-2014-nmt} in abstractive summarization and achieve promising results.
Further improvements are brought with recurrent decoders~\cite{N16-1012}, selective gate network~\cite{P17-1101}, abstract meaning representation~\cite{takase2016neural}, hierarchical networks~\cite{K16-1028} and variational auto-encoders~\cite{D16-1031}.
In terms of QFS, Nema et al.~\cite{nema2017diversity} introduce a query attention model
in the encoder-decoder framework, and a diversity attention model to alleviate the problem of repeating phrases in summary. 
Query relevance, multi-document coverage, and summary length constraints are incorporated into seq2seq models to improve QFS performance~\cite{baumel2018query}.
Most QFS work involves long natural language questions as the queries, while we focus on short search queries in this paper.

\noindent \textbf{Constrained Sentence Generation.} Constrained seq2seq sentence generation, considering external information during generation, are widely studied in human-computer conversation systems and e-commerce applications. 
Mou et al.~\citep{mou2016sequence} propose a content-introducing approach to dialogue systems, which can generate a reply containing the given keyword.
Yao et al.~\citep{yao-etal-2017-towards} propose an implicit content-introducing method that incorporates additional information into the seq2seq model via a hierarchical gated fusion unit. 
Xing et al.~\citep{xing2017topic} consider incorporating topic information into a seq2seq framework to generate informative responses for chatbots.
Sun et al.~\citep{Sun:2018:MPN:3269206.3271722} propose a multi-source pointer network~\citep{NIPS2015_5866} by adding a new knowledge encoder to retain the key information during product title generation.
In e-commerce search scenarios, a query generation task is proposed to improve long product title compression performance in a multi-task learning framework~\citep{Wang2018AML}. 
Chen et al.~\cite{chen2019towards} propose a knowledge-based personalized (KOBE) product description generation model in the context of e-commerce which considers product aspects and user categories during text generation.  
Duan et al.~\cite{duan2020query} propose a query-variant advertisement generation model that takes keywords and associated external knowledge as input during training and adds different queries during inference.
Abstractive tip generation is first studied and deployed in recommendation systems~\cite{li2017neural}, where tip generation is jointly optimized with rating prediction using a multi-task learning manner.
Some researchers also capture the intrinsic language styles of users via variational auto-encoders to generate personalized tips~\cite{li2019neural}. 
To take the query impact into account, this paper proposes query-aware tip generation for vertical search.
We consider query information in both encoder and decoder sides to generate query-aware tips, that are intuitive but effective and of great business values in vertical search scenarios.

\section{Conclusion}
Vertical search results are devoted to a certain media type or genre of content.
Taking Dianping as an example, given a query, the vertical search engine usually returns a list of relevant POIs (i.e., restaurants) to users. 
To improve the user experience, some extra information need to be presented together with the search results.
Tip, a concise summary of genuine user reviews, is an intuitive and complementary form to help users get a quick insight into the search results. 
This paper studies the task of query-aware tip generation for vertical search.
We propose an intuitive and effective query-aware tip generation framework.
Two specific adaptations for the Transformer and the RNN architectures are developed.
Extensive experiments on both public and realistic datasets reveal the effectiveness of our proposed approach.
The online deployment experiments on Dianping demonstrate the promising business value of the query-aware tip generation framework.

\clearpage
\bibliographystyle{ACM-Reference-Format}
\bibliography{7-reference}
\end{document}